\documentclass[conference]{IEEEtran}
\IEEEoverridecommandlockouts
\usepackage{cite}
\usepackage{amsmath,amssymb,amsfonts}
\usepackage{algorithmic}
\usepackage{graphicx}
\usepackage{textcomp}
\usepackage{xcolor}
\usepackage[draft]{minted}

\usepackage{tcolorbox}
\tcbuselibrary{minted,breakable,xparse,skins}

\definecolor{bg}{gray}{0.95}

\DeclareTCBListing{mintedbox}{O{}m!O{}}{%
  breakable=true,
  listing engine=minted,
  listing only,
  minted language=#2,
  minted style=default,
  minted options={%
    linenos,
    gobble=0,
    breaklines=true,
    breakafter=,,
    fontsize=\small,
    numbersep=8pt,
    #1},
  boxsep=0pt,
  left skip=0pt,
  right skip=0pt,
  left=25pt,
  right=0pt,
  top=3pt,
  bottom=3pt,
  arc=5pt,
  leftrule=0pt,
  rightrule=0pt,
  bottomrule=2pt,
  toprule=2pt,
  colback=bg,
  colframe=orange!70,
  enhanced,
  overlay={%
    \begin{tcbclipinterior}
    \fill[orange!20!white] (frame.south west) rectangle ([xshift=20pt]frame.north west);
    \end{tcbclipinterior}},
  #3}

\def\BibTeX{{\rm B\kern-.05em{\sc i\kern-.025em b}\kern-.08em
    T\kern-.1667em\lower.7ex\hbox{E}\kern-.125emX}}
\begin{document}

\title{SQUAT: Stateful Quantization-Aware Training in Recurrent Spiking Neural Networks\\

}

\author{\IEEEauthorblockN{Sreyes Venkatesh}
\IEEEauthorblockA{\textit{ECE},
\textit{UC Santa Cruz}\\
Santa Cruz, CA, USA \\
spvenkat@ucsc.edu}
\and
\IEEEauthorblockN{Razvan Marinescu}
\IEEEauthorblockA{\textit{CSE},
\textit{UC Santa Cruz}\\
Santa Cruz, CA, USA \\
ramarine@ucsc.edu}
\and
\IEEEauthorblockN{Jason K. Eshraghian}
\IEEEauthorblockA{\textit{ECE},
\textit{UC Santa Cruz}\\
Santa Cruz, CA, USA \\
jeshragh@ucsc.edu}
}

\maketitle

\begin{abstract}
Weight quantization is used to deploy high-performance deep learning models on resource-limited hardware, enabling the use of low-precision integers for storage and computation. Spiking neural networks (SNNs) share the goal of enhancing efficiency, but adopt an `event-driven' approach to reduce the power consumption of neural network inference. While extensive research has focused on weight quantization, quantization-aware training (QAT), and their application to SNNs, the precision reduction of state variables during training has been largely overlooked, potentially diminishing inference performance. This paper introduces two QAT schemes for stateful neurons: (i) a uniform quantization strategy, an established method for weight quantization, and (ii) threshold-centered quantization, which allocates exponentially more quantization levels near the firing threshold. Our results show that increasing the density of quantization levels around the firing threshold improves accuracy across several benchmark datasets. We provide an ablation analysis of the effects of weight and state quantization, both individually and combined, and how they impact models. Our comprehensive empirical evaluation includes full precision, 8-bit, 4-bit, and 2-bit quantized SNNs, using QAT, stateful QAT (SQUAT), and post-training quantization methods. The findings indicate that the combination of QAT and SQUAT enhance performance the most, but given the choice of one or the other, QAT improves performance by the larger degree. These trends are consistent all datasets. Our methods have been made available in our Python library snnTorch: https://github.com/jeshraghian/snntorch.

\end{abstract}

\section{Introduction}

The development of low-power neural networks  is crucial for enabling operation on portable and edge devices \cite{7966166, 7331375, Kopuklu_2019_ICCV}. Techniques such as pruning~\cite{janowsky1989pruning}, specialized data encoding~\cite{auge2021survey}, model compression~\cite{cheng2018model}, early exiting~\cite{teerapittayanon2016branchynet}, amongst many others, can be used to reduce the computational cost of running a neural network~\cite{han2015deep, He_2017_ICCV}. In tandem with all these approaches, most edge devices require low or fixed precision model parameters. Quantized neural networks (QNNs) require full precision weights to be approximated down to lower-capacity representations which further reduces memory and computation demands~\cite{hubara2016quantized, zhou2017balanced}. 

Spiking neural networks (SNNs), drawing from models of biological neurons, use binarized activations, enabling a neuron to either emit a spike or remain inactive. This binary representation allows neuromorphic processors to bypass certain computations and memory accesses typical in conventional deep learning, leading to significant power savings~\cite{pfeiffer2018deep, azghadi2020hardware, roy2019towards, schmidgall2023brain}. The hidden state of a spiking neuron is often governed by a dynamical system, and is thought to encode information which is then communicated through the firing pattern of the neuron. These patterns can vary in spike timing, frequency, intervals between spikes, amongst many other theories, offering a range of encoding possibilities~\cite{petro2019selection, nath2024optically, boahen2022dendrocentric, eshraghian2018neuromorphic}.

SNNs are often tailored for edge devices and have demonstrated powerful computational capabilities even with considerably small models~\cite{shrestha2018slayer, henkes2022spiking, yang2023neuromorphic, barchid2023spiking}, and larger-scale models are also emerging~\cite{zhu2023spikegpt}. Fixed-precision representations in SNN accelerators are standard, and quantized SNNs (QSNNs) are commonplace when deploying spike-based models on neuromorphic hardware~\cite{davies2018loihi, orchard2021efficient, bos2023sub, richter2023speck, pedersen2023neuromorphic, rahimi2020complementary}. 

By default, the most widely used deep learning Python libraries train models with full precision parameters. Quantizing a model after it has been trained in full precision is known as `Post-Training Quantization' (PTQ). Performance can be enhanced further by using `Quantization-Aware Training' (QAT), where weights are quantized during the forward-pass but gradients are calculated in full precision. The quantization step is ignored during error backpropagation as it is a non-differentiable function. This process allows for the loss from the model to account for truncation errors made during quantization. This modification to the backpropagation process takes more computational resources, but enhances accuracy.

To improve loss convergence, training QNNs and QSNNS both benefit from modifying the training process~\cite{8259423, 9586323, neftci2017event, shrestha2019approximating, esser2015backpropagation,6889876}.
In general, noise has been regarded as an obstacle towards convergence\footnote{Note that targeted non-systematic noise has been demonstrated to assist in convergence, due to the noise acting as a regularizer \cite{8803055, 548170, noh2017regularizing}, and is not included in this claim} \cite{dundar1995effects, 964997}. Though it was suggested in Ref.~\cite{eshraghian2022navigating} that QSNNs are tolerant to truncation, provided that any rounding errors do not trigger a threshold-crossing of the neuron state, thus either hallucinating or eliminating a spike~\cite{eshraghian2022memristor}.

Most advances in low-precision neural networks apply variations of QAT to learning fixed-precision weights~\cite{8325325, lui2021hessian, chowdhury2021spatiotemporal}. At one extreme, Ref.~\cite{ma2024era} demonstrated promising performance in a ternary language model with over one billion weights. These variants of QAT may involve alternative distributions from which quantization levels are sampled from (e.g., uniform quantization, exponentially distributed quantization). The most common practice is to rely on full precision simulators that emulate quantized weights. This is done by restricting the permitted full precision levels of weights, and then training such models with QAT (e.g., using the Brevitas Python library~\cite{brevitas}). 

While it may be conceptually trivial to extend this practice to the hidden state of neurons: e.g., quantizing states during the forward-pass of training such that the loss accounts for truncation error of states, it is not done in practice. 
There are a variety of reasons why this is the case: i) an absence of tools that simplify quantization of states; ii) applying QAT to neuron states is computationally expensive: weights are constant over sequence steps, whereas the state must be re-quantized at every sequence step; iii) modern deep learning is less reliant on stateful and sequential neural networks, and iv) there are typically more weights than there are neurons, so optimizing for weights may be thought of as more important $-$ i.e., it is just `easier' to apply post-quantization to the states once a QSNN is deployed.

This study analyzes two forms of stateful quantization aware training (`SQUAT') for training QSNNs: `uniform quantization' and `exponential quantization'. Exponential quantization allocates more states about the firing threshold of a neuron as near-threshold activity of a neuron is likely to have a greater impact on downstream neurons. Therefore, higher precision is more useful at this point of criticality.

Beyond proposing SQUAT, we present an empirical analysis evaluating how the quantization of weights and states compare in both the post-quantization and quantization-aware training regimes. A comprehensive evaluation is performed across three different datasets of varying difficulty and modality: an image dataset, FashionMNIST \cite{xiao2017fashionmnist}, an auditory dataset, Spiking Heidelberg Digits (SHD) \cite{9311226}, and an event-based dataset, the DVS Gesture Dataset \cite{8100264}. Specifically, we test 2-bit, 4-bit, and 8-bit QSNNs under the following cases: 

\begin{itemize}

\item  PTQ of states and weights, both together and separately, for uniform quantization
\item PTQ of states and weights, both together and separately, for exponential quantization
\item QAT and SQUAT, both together and separately, for uniform quantization
\item QAT and SQUAT, both together and separately, for exponential quantization
\item A high precision SNN baseline for each experiment
\end{itemize}

In total, we conduct 129 experiments across three trials each culminating in insights about the impacts of quantization across weights and states, and how to gain the `last mile' of performance from an SNN. Our findings are summarized below:

\begin{itemize}
    \item 8-bit QSNNs perform competitively against their full precision counter parts for all tested quantization schemes: i.e., PTQ vs. QAT and Exponential Distribution vs Uniform Distribution. 
    \item 4-bit and 2-bit models are far more sensitive to quantization and require more care during training. 
    \item QAT (weights) and SQUAT (states) combined consistently provided the best performance across all bit-widths for fixed-precision performance.
    \item When using either SQUAT or PTQ, exponentially distributed levels centered about the threshold consistently improved accuracy for all experiments.
    \item QAT-only consistently outperforms SQUAT-only. Given a large computational budget, combining both is most effective. Where there are limited training resources and one technique must be prioritized, QAT is more important than SQUAT. 
\end{itemize}

Finally, we release code to assist other researchers to train QSNNs using SQUAT in a straightforward and intuitive manner.

\section{Background}

\subsection{Spiking Neuron Model}

The spiking neuron model adopted is the leaky integrate-and-fire neuron:

\begin{equation}
\tau_{m}\frac{dU}{dt}=-U+RI,
\end{equation}

\noindent where $\tau_{m}$ is the time constant of membrane potential, {\itshape U} is the membrane potential,  {\itshape R} is the passive resistance of the membrane of the neuron, and {\itshape I} is the input current \cite{hunsberger2015spiking}. The above equation when governed by discrete time dynamics and represented in a recurrent manner can be represented by:


\begin{equation}
  u_{t+1}^{j}=\beta u_{t+1}^{j} + \sum_{i} w^{ij} z_t^{j} - z_t^{j}\theta 
\end{equation}

\begin{equation}
  z_t^{j} = \begin{cases}
        1, & \text{if } u_{t+1}^{j} > \theta\\
        0, & \text{otherwise } 
    \end{cases}
\end{equation}

\noindent where $u_{t+1}^{j}$ is the membrane potential of neuron {\itshape j} at time {\itshape t}; $\beta$ is the inverse time constant of membrane potential; $w^{ij}$ is the synaptic weight between neurons i and j. The neuron is reset by the threshold $\theta$ every time a spike $z_t^{j}$ is emitted. Eq.~(2) shows how a spike emitted at each instance of time where the membrane potential exceeds the firing threshold.

\begin{figure*}[ht]
    \centering
    \includegraphics[scale=0.8]{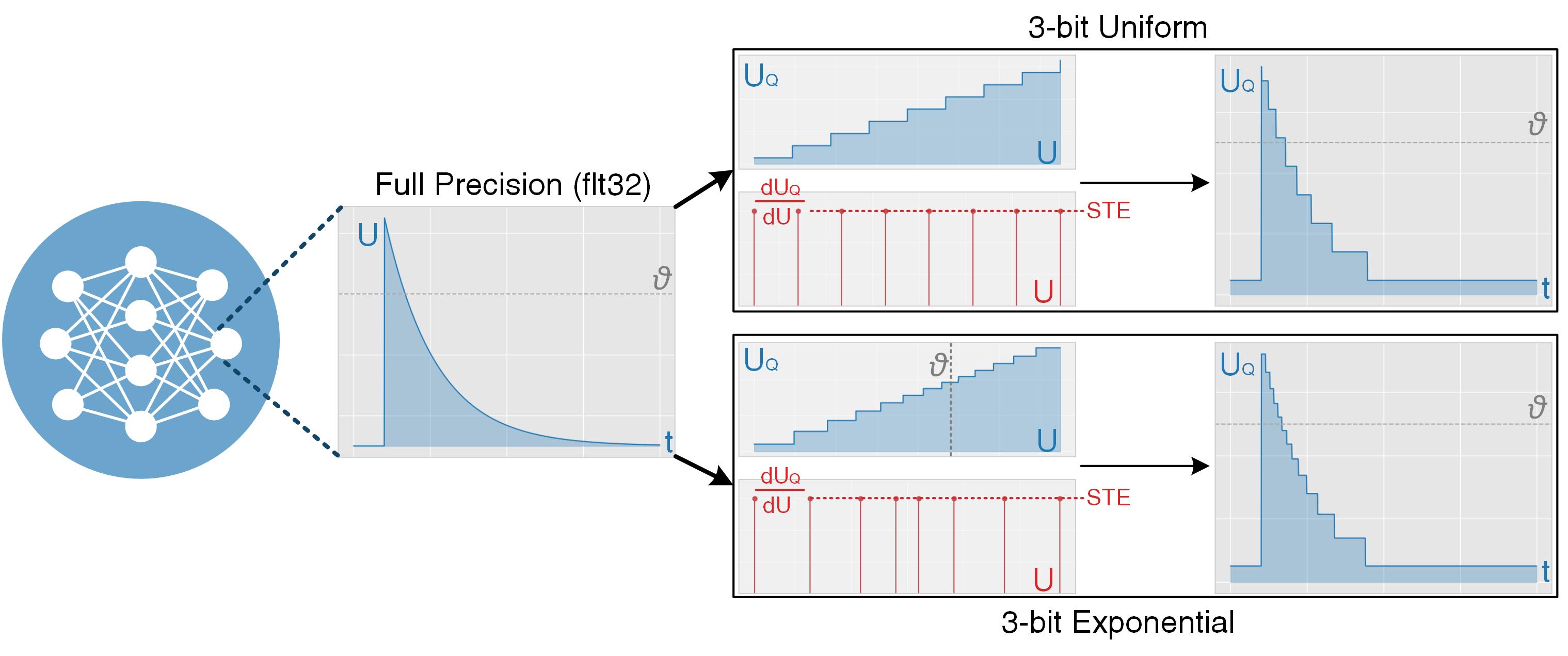}
    \caption{A graphical depiction of stateful quantization. On the left, a membrane potential trajectory is depicted in full precision. 
            The state can be quantized either via uniform or exponential quantization. A 3-bit (8 levels) quantization scheme is illustrated. 
            In uniform quantization, the permissible levels are evenly distributed. In exponential quantization, the permissible levels are closer about the threshold, and widely distributed moving further away from the threshold. A `straight through estimator' (STE) is also depicted to address the non-differentiability of quantization.
            }
    \label{fig:abstract}
\end{figure*}

\subsection{Hard Thresholds in QSNNs}
There are two non-differentiable operations in SNNs: one is from spike generation (Eq. (3)), and another is applied to the weights during QAT and to the states during SQUAT. Addressing these challenges is quite well-established:

\begin{itemize}
    \item \textbf{Spike Non-Differentiability:} The non-differentiability of spikes has been addressed by applying a step function during the forward-pass as per (3), and smoothing it out into a `surrogate gradient' in the backward-pass~\cite{neftci2019surrogate}.
    \item \textbf{QAT:} QAT follows a similar approach, though with additional steps as the weight is a learnable parameter:
    \begin{enumerate}
        \item Quantized weights are used for computation during the forward-pass, while the original full precision representations are stored in memory.
        \item A loss is calculated based on quantized weights.
        \item Gradients are calculated while neglecting the gradient of the quantization operator.
        \item Weight updates are applied to the full precision weights rather than the quantized weights.
        \item The process is repeated.
    \end{enumerate}

\end{itemize}

The quantization operator is neglected during the gradient calculation as it is non-differentiable and would otherwise null the gradient signal. To remedy this, a `straight through estimator' (STE) is used to bypass the threshold operator during QAT. The STE acts as an approximate gradient that smooths the thresholding function during training \cite{bengio2013estimating, yin2019understanding, shekhovtsov2021reintroducing, liu2022nonuniformtouniform, fan2022training}. More formally, the surrogate gradient of the quantized weight $w_q$ with respect to the real weight $w_r$ is:

\begin{equation}
\frac{\partial w_q}{\partial w_r} = 1
\end{equation}

Alternatively, PTQ can be applied to a pre-trained model where weights are quantized after training a full precision model~\cite{Putra_2021}. 
Whilst computationally cheaper, it also leads to a drop in model performance. 

SQUAT extends these principles by calculating a loss that is aware of the quantization of states, while bypassing the non-differentiability using a STE. The next section provides further detail on the variants of SQUAT that we propose and test.

\section{Methods}

\subsection{Stateful Quantization-Aware Training}

For {\itshape{n}}-bit quantization, the number of permitted levels $Q_l$ with respect to the number of bits allocated $n$ is $Q_l = 2^n$.
 We implement SQUAT by distributing these $Q_l$ levels using two different methods, both of which are illustrated in Fig.~1:

\begin{itemize}
    \item \textbf{Uniform Quantization: } The permissible quantization levels are uniformly distributed between the minimum $U_{\rm min}$ and maximum $U_{\rm max}$ membrane potentials. $U_{\rm min}$ and $U_{\rm max}$ are recalculated for each forward-pass.
    \item \textbf{Exponential Quantization:} Similar to uniform quantization, the maximum and minimum possible voltages for the membrane potential are applied as the upper and lower bound of permissible levels. However, exponentially more quantization levels are allocated for the threshold. In deep learning, exponentially distributed weights are applied about `0'. However, spiking neurons only communicate with one another when the membrane potential reaches the threshold. Intuitively, it stands to reason that a neuron requires more precision about the threshold as this is the regime where network activity is determined.
\end{itemize}

\subsubsection{Uniform SQUAT}
More formally, in uniform quantization, we divide the range between the minimum (\( U_{\text{min}} \)) and maximum (\( U_{\text{max}} \)) membrane potentials into equal intervals. Let's denote the number of intervals as \( N = 2^n - 1 \). The quantized value \( U_q \) can be calculated as follows:

\[
U_q = U_{\text{min}} + \Bigg[ \frac{U - U_{\text{min}}}{\Delta U} \Bigg] \cdot \Delta U
\]

\noindent where \( \Delta U = \frac{U_{\text{max}} - U_{\text{min}}}{N} \) is the size of each interval. The operator $[ \cdot ]$ rounds the argument to the nearest integer. It is assumed that $U_{\rm min}$ and $U_{\rm max}$ have been scaled and offset such that $U_{\rm min}=0$ and $U_{\rm max}=N$.

\subsubsection{Exponential SQUAT}
In this case, the quantization becomes finer as it approaches the threshold \( U_{\text{th}} \). The quantization is described using a pair of exponential functions, one for values below the threshold and another for values above it, while being clipped at \( U_{\text{min}} \) and \( U_{\text{max}} \). The quantized value \( U_q \) can be expressed as:

\[
U_q = 
\begin{cases} 
U_{\text{min}} + \Delta U \cdot \left\lfloor \frac{1 - e^{-a(U - U_{\text{min}})}}{\Delta U} \right\rfloor, & \text{if } U <\theta \\
U_{\text{max}} - \Delta U \cdot \left\lfloor \frac{1 - e^{-b(U_{\text{max}} - U)}}{\Delta U} \right\rfloor, & \text{if } U \geq\theta
\end{cases}
\]

\noindent where \( a \) and \( b \) are the exponents that determine the steepness of the exponential curves below and above the threshold, respectively.
This formulation allows for exponential quantization of the membrane potential while considering the resolution implied by the number of bits \( N \).

\subsubsection{Straight-Through-Estimator}
As with the STE from (4), the same is applied to the membrane potential:

\begin{equation}
    \frac{\partial U_q}{\partial U} = 1
\end{equation}

These equations provide a mathematical framework for quantizing the membrane potential in both uniform and exponentially varying manners near the threshold. 

\subsection{Testing}
We assess performance across three data sets: FashionMNIST, Spiking Heidelberg, and DVS Gesture. For each dataset:

\begin{itemize}
    \item A full precision baseline is obtained on a lightweight architecture.
    \item A PTQ baseline is obtained by converting the full precision baseline to 8-bits, 4-bits, and 2-bits for weights-only, states-only, and both. For the case where- only the weights are quantized, the states are left in full-precision and vice versa. 
    \item QAT-only results are obtained by retraining the model across 8-bit, 4-bit, and 2-bit weights. States are left in full precision.
    \item SQUAT-only results are obtained by retraining the model across 8-bit, 4-bit, and 2-bit weights, for uniform quantization and exponential quantization. Weights are left in full precision.
    \item QAT and SQUAT are both applied across 8-bit, 4-bit, and 2-bit weights, for uniform quantization and exponential quantization

\end{itemize}




 \subsection{Training}

For all experiments, a threshold-shifted arc-tangent surrogate gradient is used to deal with the spike non-differentiability:

\begin{equation}
\frac{\delta z}{\delta U} = \frac{1}{\pi} \frac{1}{[1 + (\pi U\alpha)^2]}
\end{equation}

Each network is tested with a cosine annealing learning rate scheduler and the Adam Optimizer~\cite{he2018bag, loshchilov2017sgdr}.
To maintain the hardware benefits obtained from operating on lower-bit variables, we intentionally expose the network to a discontinuous loss landscape with flat surfaces and use the more challenging task of using the total spike count per neuron as the logits. Many state-of-the-art results opt to use a `read-out' layer instead, where a standard artificial neuron node is applied at the final layer instead of a spiking layer.  

For the DVS-Gesture dataset we use the Mean-Squared Error loss with respect to the spikes of each output neuron: $z_t^{j}$ and the target spike count $c^{j}$. 
Each of those losses is then summed over $M$ output classes.

\begin{equation}
\mathcal{L}_{MSE}= \sum_{j}^{M} \sum_t (c^{j} -  z_t^{j})^2,
\end{equation}

For the FashionMNIST and Spiking Heidelberg Digits datasets we use the cross entropy loss as applied to the spike count:

\begin{equation}
\mathcal{L}_{CE} = \frac{1}{C} \sum_{j}^{M} \sum_{t=0}^{C} N (log(p^{j}[t]), Y^{j})
\end{equation}

\noindent where $C$ is the number of time steps, $p^{j}[t]$ is the softmax probability of the spike count of the output neuron {\itshape{j}} at time step {\itshape{t}}, $N$ is the negative log likelihood loss function, and $Y^j$ is the target spike count of output neuron {\itshape{j} }.    

\subsection{Model Architecture}

Given the notation $C_{\rm  out}$Conv$k$ and $N_{\rm in}$Dense$N_{\rm out}$, where $C_{\rm out}$ is the number of output channels, $k$ is the kernel dimension, and $N_{\rm in}$ and $N_{\rm out}$ are the input-output dimensions of a linear layer, the model architectures used were: 
\begin{itemize}
    \item FashionMNIST: 16Conv5-MP2-64Conv5-MP2-1024Dense10. Batch normalization is applied to all convolutional layers~\cite{srivastava2014dropout}.
    \item DVS Gesture: 16Conv5-MP2-32Conv5-MP2-8800Dense11. Dropout is applied during training. Batch normalization is applied to all convolutional layers.
    \item Spiking Heidelberg Digits: 700Dense1000-1000Dense20. Each linear layer is followed by batch normalization operation. Dropout is also applied.
\end{itemize}

\section{Experimental Results}  
Brevitas was used to apply uniformly quantize the network weight and bias parameters during training and testing; snnTorch was used to instantiate the SNNs, and PyTorch 2.0 for training and testing. Each result shown is the average taken across three trials.

The FashionMNIST and SHD dataset simulations were limited to 100 epochs with early stopping of 20 epochs applied. The DVS Gesture Dataset accuracy was obtained by running 500 epochs. More details are given in the individual sections of the datasets. 

\begin{table}[ht]
\centering
\caption{Full Precision Accuracy}
\begin{tabular}{c|c}
            Dataset & Accuracy           \\ \hline
            FMNIST        & 90.87      \\
            SHD       & 79.78         \\
            DVS  & 86.24        \\

\end{tabular}
\end{table}

\subsection{FashionMnist}

\begin{figure*}[ht]
    \centering
    \includegraphics[scale=0.425]{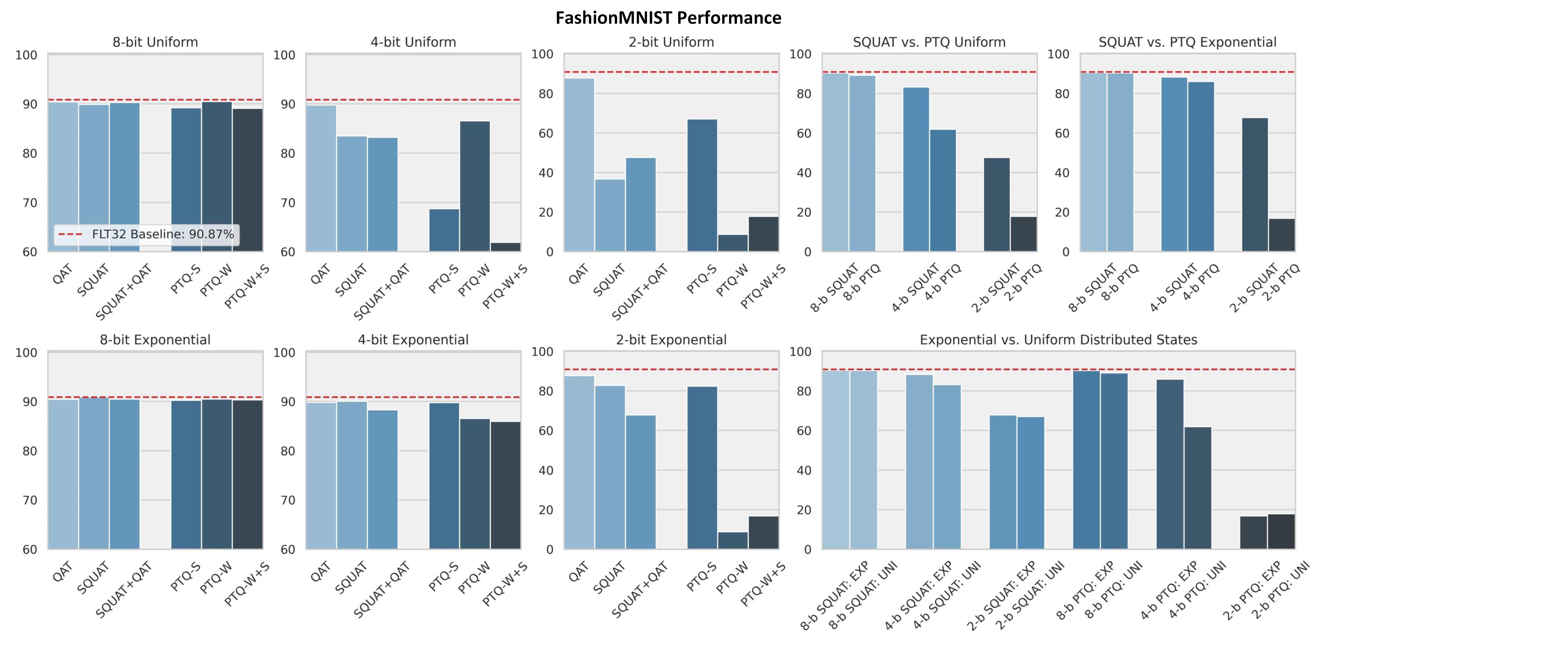}
    \caption{FashionMNIST performance. Top row: (i) 8/4/2-b uniformly distributed states across QAT (N-b weights, flt32 states), SQUAT (flt32 weights, N-b states), SQUAT+QAT (N-b weights, N-b states), and PTQ-S (flt32 weights, N-b states), PTQ-W (N-b weights, flt32 states), PTQ-W+S (N-b weights, N-b states). (ii) SQUAT vs PTQ uniformly and exponentially distributed states (SQUAT and QAT are both applied across N-b weights and states, and compared against PTQ of N-b states and weights). Bottom row:  
    (iii) 8/4/2-b exponentially distributed states across QAT (N-b weights, flt32 states), SQUAT (flt32 weights, N-b states), SQUAT+QAT (N-b weights, N-b states), and PTQ-S (flt32 weights, N-b states), PTQ-W (N-b weights, flt32 states), PTQ-W+S (N-b weights, N-b states). (iv) Comparison between exponential and uniformly distributed states: SQUAT+QAT are used across N-b states and weights, then N-b PTQ is used across N-b states and weights.
            }
    \label{fig:fmnist}
\end{figure*}

The FashionMNIST dataset contains ten classes of clothing items and accessories \cite{xiao2017fashionmnist}. The raw FashionMNIST dataset was repeatedly passed to the network for 25 time steps of simulation without encoding.

Table II displays the results of exponentially-distributed states, Table III displays the results of uniformly-distributed states, and Fig.~2 shows barplots of the same results organized to provide easier visual comparisons. Quantizing weights to 8-bits had negligible impact on accuracy, regardless of the quantization configuration (SQUAT/QAT or PTQ).
At 4-bits, exponentially-distributed states using SQUAT significantly outperforms uniformly-distributed states with SQUAT (Table III) in all test cases by a significant margin. 
Finally, 2-bit weights destroy performance when using PTQ, but the model can be salvaged by combining QAT and SQUAT using exponentially distributed levels centered about the threshold (i.e., an accuracy boost from 16.81\% to 67.82\%).

Note PTQ and QAT of weights are the same between uniform and threshold-centered quantization, as quantization levels is only applied to states, and these two configurations do not quantize the states. The same is true for all datasets, but are included for ease of reference.


%
%
\begin{table}[ht]
\caption{FMNIST Exponentially Centered Quantized Performance}
\begin{tabular}{c|ccc}
                  & 8-bit      & 4-bit   & 2-bit       \\ \hline
QAT States        & 90.87      &  90.04  &  82.79\\
QAT Weights       & 90.43 & 89.79 & 87.77       \\
QAT Weights and States  & 90.49 & 88.29 & 67.82       \\
PTQ States        & 90.26     & 89.78  & 82.36\\ 
PTQ Weights       & 90.48     & 86.56  & 8.72 \\
PTQ Weights and States & 90.32 & 85.95 & 16.81\\

\end{tabular}
\end{table}

\begin{table}[ht]
\caption{FMNIST Uniform Quantized Performance}
\begin{tabular}{c|ccc}
                  & 8-bit     & 4-bit   & 2-bit       \\ \hline
QAT States        & 89.91     & 83.50   & 36.76\\
QAT Weights       & 90.43 & 89.79 & 87.77       \\
QAT Weights and States  & 90.31 & 83.21 & 47.58       \\
PTQ States        & 89.22     & 68.72  & 67.06\\ 
PTQ Weights       & 90.48     & 86.56  & 8.72 \\
PTQ Weights and States & 89.11 & 61.87 & 17.87 \\

\end{tabular}
\end{table}

\subsection{Spiking Heidelberg Digits}

\begin{figure*}[ht]
    \centering
    \includegraphics[scale=0.425]{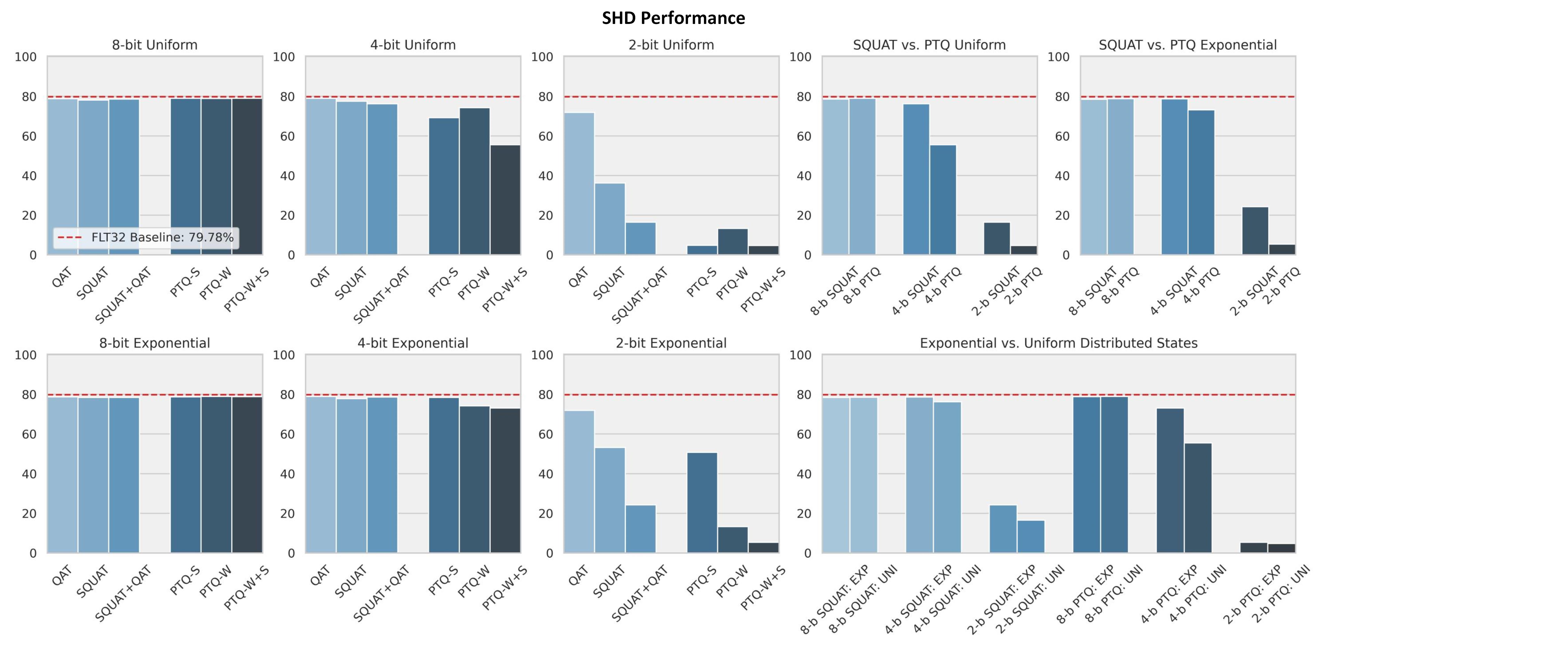}
    \caption{SHD performance. Top row: (i) 8/4/2-b uniformly distributed states across QAT (N-b weights, flt32 states), SQUAT (flt32 weights, N-b states), SQUAT+QAT (N-b weights, N-b states), and PTQ-S (flt32 weights, N-b states), PTQ-W (N-b weights, flt32 states), PTQ-W+S (N-b weights, N-b states). (ii) SQUAT vs PTQ uniformly and exponentially distributed states (SQUAT and QAT are both applied across N-b weights and states, and compared against PTQ of N-b states and weights). Bottom row:  
    (iii) 8/4/2-b exponentially distributed states across QAT (N-b weights, flt32 states), SQUAT (flt32 weights, N-b states), SQUAT+QAT (N-b weights, N-b states), and PTQ-S (flt32 weights, N-b states), PTQ-W (N-b weights, flt32 states), PTQ-W+S (N-b weights, N-b states). (iv) Comparison between exponential and uniformly distributed states: SQUAT+QAT are used across N-b states and weights, then N-b PTQ is used across N-b states and weights.
            }
    \label{fig:shd}
\end{figure*}

The Spiking Heidelberg Digits is an audio based classification dataset of 20 possible output classes. There are approximately 10,000 recordings of spoken digits from 0 to 9 in both English and German \cite{9311226}.  

The SHD dataset was passed to the network for 25 time steps during training and 30 time steps during testing. Table IV displays the average test accuracy over 100 epochs across 3 trials each for exponentially-distributed states, Table V displays the same data for uniformly-distributed states, and Fig.~3 illustrates the same data as a barplot. 

Similarly to the FashionMNIST dataset, 8-bit quantization had little to no effect on accuracy, regardless of the quantization configuration. At 4-bits, exponentially-distributed quantization of states improved performance relative to PTQ. At 2-bits, there is significant variation across all configurations. At this point, when both states and weights are quantized down to 2-bits, these models are quite unstable. But using QAT and SQUAT together nonetheless outperforms PTQ methods by a significant margin. Exponentially distributed state levels outperforms uniformly distributed levels by approximately 8\% at 2-bits.

\begin{table}[ht]
\caption{SHD Exponentially Centered Quantized Performances}
\begin{tabular}{c|ccc}
                  & 8-bit      & 4-bit   & 2-bit       \\ \hline
QAT States        & 78.48      &  77.81  &  53.25\\
QAT Weights       & 78.72      & 78.98   & 71.88      \\
QAT Weights and States  & 78.46 & 78.63 & 24.26       \\
PTQ States        & 78.76     & 78.42  & 50.70\\ 
PTQ Weights       & 78.89     & 74.19  & 13.27 \\
PTQ Weights and States & 78.79 & 73.10 & 5.33\\

\end{tabular}
\end{table}

\begin{table}[ht]
\caption{SHD Uniform Centered Quantized Performance}
\begin{tabular}{c|ccc}
                  & 8-bit     & 4-bit   & 2-bit       \\ \hline
QAT States        & 78.03      &  77.43  &  36.27\\
QAT Weights       & 78.72 & 78.98 & 71.88       \\
QAT Weights and States  & 78.52 & 76.23 & 16.48       \\
PTQ States        & 78.96     & 69.16  & 4.74\\ 
PTQ Weights       & 78.89     & 74.19  & 13.27 \\
PTQ Weights and States & 78.94  & 55.47 & 4.69 \\

\end{tabular}
\end{table}

\subsection{DVS Gesture}

\begin{figure*}[ht]
    \centering
    \includegraphics[scale=0.425]{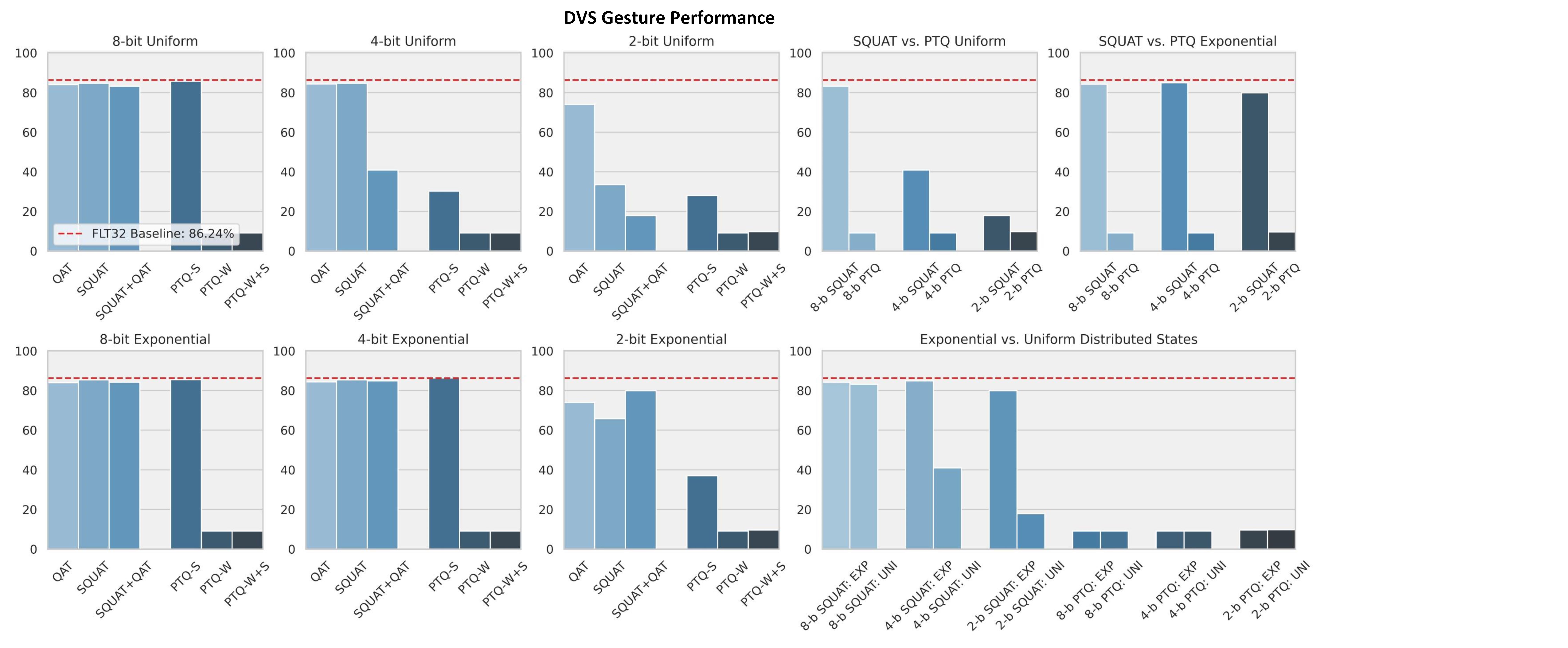}
    \caption{DVS Gesture Dataset performance. Top row: (i) 8/4/2-b uniformly distributed states across QAT (N-b weights, flt32 states), SQUAT (flt32 weights, N-b states), SQUAT+QAT (N-b weights, N-b states), and PTQ-S (flt32 weights, N-b states), PTQ-W (N-b weights, flt32 states), PTQ-W+S (N-b weights, N-b states). (ii) SQUAT vs PTQ uniformly and exponentially distributed states (SQUAT and QAT are both applied across N-b weights and states, and compared against PTQ of N-b states and weights). Bottom row:  
    (iii) 8/4/2-b exponentially distributed states across QAT (N-b weights, flt32 states), SQUAT (flt32 weights, N-b states), SQUAT+QAT (N-b weights, N-b states), and PTQ-S (flt32 weights, N-b states), PTQ-W (N-b weights, flt32 states), PTQ-W+S (N-b weights, N-b states). (iv) Comparison between exponential and uniformly distributed states: SQUAT+QAT are used across N-b states and weights, then N-b PTQ is used across N-b states and weights.
            }
    \label{fig:dvs}
\end{figure*}

The DVS Gesture dataset is an event based dataset with 11 output classes consisting of various hand gestures~\cite{8100264}. The DVS dataset was passed to the network for 25 time steps during training and 150 time steps during testing. Of all the three datasets, the DVS gesture dataset was the most sensitive to quantization. This may be because it has the largest number of neurons and synapses in its architecture, and thus, the most truncation error accumulated in the loss. With reference to Table VI for exponentially distributed quantizations, Table VII for uniformly distributed levels, and Fig.~4 for the corresponding barplots, all PTQ training runs at 2-bits yielded rather hopeless models despite our best hyperparamter sweep efforts. However, the combination of exponentially-distributed levels, together with QAT and SQUAT boosted performance back up to 79.89\%, as against 9.59\% using PTQ (i.e., random chance), and 28.03\% when using uniform distributed SQUAT + QAT. This highlights the significant importance of adopting better distribution strategies in the extreme quantization regime. The trends for all three datasets remain incredibly consistent, with huge performance margins for exponentially-distributed hidden state levels in the extreme quantization regime. 

\begin{table}[ht]
\caption{DVS Exponentially Centered Quantized Performance}

\begin{tabular}{c|ccc}
                  & 8-bit      & 4-bit   & 2-bit       \\ \hline
QAT States        & 85.35 & 85.35 & 65.78\\
QAT Weights       & 83.97 & 84.34 & 73.99       \\
QAT Weights and States  & 84.22 & 84.85 & 79.89       \\ 
PTQ States        & 85.48 & 86.36 & 36.95\\ 
PTQ Weights       & 9.09 & 9.09 & 9.09 \\
PTQ Weights and States & 9.09 & 9.09 & 9.59\\

\end{tabular}
\end{table}

\begin{table}[ht]
\caption{DVS Uniform Centered Quantized Performance}
\begin{tabular}{c|ccc}
                  & 8-bit     & 4-bit   & 2-bit       \\ \hline
QAT States        & 84.72 & 84.72 & 33.46\\
QAT Weights       & 83.97 & 84.34 & 73.99       \\
QAT Weights and States & 83.20 & 40.91 & 17.80       \\
PTQ States        & 85.73 & 30.17 & 28.03 \\ 
PTQ Weights       & 9.09 & 9.09 & 9.09 \\
PTQ Weights and States & 9.09 & 9.09 & 9.72 \\

\end{tabular}
\end{table}

\section{Discussion}

\textbf{Deducing the optimal quantization scheme:} For the 8-bit experiments one can see that the performance of the model was independent of the given quantization scheme, which is generally quite unsurprising~\cite{nahshan2021loss}. Whereas for the 4-bit and 2-bit scenarios, it becomes increasingly clear that the quantization method (i.e., PTQ/QAT/SQUAT/exponential/uniform-distribution) is a major factor on the performance of the network. Overall, performing QAT while also implementing exponentially-distributed quantization about the threshold on the hidden states of the neurons results in the least degradation of accuracy across low-bit quantization schemata. This is in line with our other findings of exponentially distributed levels outperforming uniformly distributed states, along with QAT performing better than PTQ.

\textbf {Membrane potentials were reaching much higher values than previously anticipated:} This strongly affected the considerations that had to be taken when optimizing the quantization of membrane potentials. A specific example being that: it was expected that the reset mechanism would have a strong influence on what the membrane potential is after it exceeds the threshold. In practice, the optimal thresholds tended to be between $0.5 < \theta < 1$, whereas the membrane potential was reaching values as high as $20$. Based on (2), it is clear that the reset mechanism was having a negligible effect on the membrane potential. Analyzing the membrane potential traces meant an alternative approach to clipping was required.

\textbf{Clipping had a significant impact on the accuracy:} Quantization sets maximum and minimum bounds on the membrane potential. Originally, it was believed that since a spike is determined by the value of membrane potential with respect to the threshold, it would be unnecessary to store membrane potentials that exceed approximately twice the threshold based on (1). However, we found that clipping must encompass the maximum and minimum membrane potential values that would be reached during the full precision training process to optimize performance. This implies that allowing membrane potentials to reach larger values allows the network to better mitigate noise. Further investigation may allocate less levels to negative hidden states, and act more closely to rectification units as these values do not contribute to spiking activity.

\textbf{Threshold centering can mitigate a significant loss of accuracy at lower bits:} For 8-bit quantization there was no significant difference as to whether the quantization was uniform across the range of membrane potentials or whether the quantization allocated more bits for values centered around the threshold. This implies that the difference in membrane potentials between time steps was larger than the smallest difference between the allocated state-levels. However for 4-bit quantization, threshold centering performed nearly as well as 8-bit across all datasets, whereas for 4-bit uniform quantization of states, accuracy becomes significantly worse as compared to 8-bit uniform quantization. The variation of accuracy for 4-bit uniform quantization was also much larger than the variation of accuracy for 8-bit uniform quantization. 


\textbf{Future Asymmetric Quantizations:} One can see that neuron states are more robust to quantization during PTQ whereas network weights are more adaptable to quantization during QAT. This aligns with the knowledge that noise from the quantization would be somewhat absorbed by the threshold dynamics of the neuron. Whereas it is clear that noise brought in to the weights after the network is finished training often leads the algorithm to poor performance. Allocating additional memory on certain parts of the network, for example, 3-bits for weights and 5- bits for states with QAT or 6-bits for weights and 2-bits for states, would allow the user to maximize the networks capacity for quantization and performance.

\section{Conclusion}

We have demonstrated the necessity of adopting both QAT and SQUAT in enabling extreme-quantization regimes of QSNNs to maintain some semblance of reasonable performance. Our findings clearly indicate that threshold centering of exponentially distributed states is better at mitigating the performance degradation that comes with lower precision. We have also demonstrated the differences in performance for QAT and PQT occur at lower bit (ie. 4-bit and 2-bit) situations. Enabling stateful quantization-aware training is effectively as simple as passing a single argument to a leaky integrate-and-fire neuron in snnTorch, simplifying the overall process of obtaining additional accuracy. Both uniform and exponential quantization schemes are available and paramterizable. Code snippets demonstrating how to use it are provided.  

\newpage

\begin{mintedbox}{python}
import torch
import snntorch as snn
from snntorch.functional import quant

# neuron parameters
beta = 0.5
thr = 5

# random input
rand_input = torch.rand(1)

# uniform quantization
q_uni = quant.state_quant(num_bits=4, uniform=True, threshold=thr)
lif_1 = snn.Leaky(beta=beta, threshold=thr, state_quant=q_uni)

# forward-pass for one step
spk, mem = lif(rand_input, mem)

# exponential quantization
q_exp = quant.state_quant(num_bits=4, uniform=False, threshold=thr)
lif_2 = snn.Leaky(beta=beta, threshold=thr, state_quant=q_exp)

\end{mintedbox}

\bibliographystyle{unsrt}
\bibliography{references}

\end{document}